# Bayesian Hierarchical Words Representation Learning

Oren Barkan[1]*, Idan Rejwan[12]*, Avi Caciularu[13]* and Noam Koenigstein[14]

[1]Microsoft,
[2]School of Computer Science, Tel Aviv University,
[3]Computer Science Department, Bar-Ilan University,
[4]Department of Industrial Engineering, Tel Aviv University
orenb@microsoft.com, {idanrejwan91, avi.c33}@gmail.com,
noamk@tauex.tau.ac.il

## Abstract

This paper presents the Bayesian Hierarchical Words Representation (BHWR) learning algorithm. BHWR facilitates Variational Bayes word representation learning combined with semantic taxonomy modeling via hierarchical priors. By propagating relevant information between related words, BHWR utilizes the taxonomy to improve the quality of such representations. Evaluation of several linguistic datasets demonstrates the advantages of BHWR over suitable alternatives that facilitate Bayesian modeling with or without semantic priors. Finally, we further show that BHWR produces better representations for rare words.

## 1 Introduction

In the last decade, a plethora of methods were proposed for learning vector representations for words (Mikolov et al., 2013; Pennington et al., 2014; Barkan, 2017), sentences (Lin et al, 2017; Barkan et al., 2020a), items (Barkan et al., 2016; Barkan et al., 2019; Barkan et al., 2020b; Barkan et al., 2020c), and medical concepts (Luo, el al., 2019). In the domain of natural language understanding, neural word embedding models are designed to learn distributed word representations as vectors in a latent space. In this space, arithmetic operations between the word vectors encode semantic and syntactic information. Specifically, the seminal works by (Mikolov et al., 2013; Pennington et al., 2014; Bojanowski et al., 2017), exhibited state-of-the-art performance on various linguistic tasks (Finkelstein et al., 2001; Luong et al., 2013; Rogers et al., 2018).

The major focus of most previous models was on optimizing the utilization of co-occurrence relations for learning representations, e.g., learning the probability of word $x$ to appear in the vicinity of word $y$. Yet, often, additional side information can be leveraged for learning finer embeddings. In this work, we focus on incorporating word semantic taxonomy, which is particularly useful for learning representations of rare words and for learning word representations from a small-size corpus.

To this end, we introduce the Bayesian Hierarchical Words Representation (BHWR) learning algorithm. BHWR presents two complementary properties: Bayesian modeling of word representations aside with hierarchical priors that naturally support semantic taxonomy. BHWR is based on a Variational Bayes (VB) optimization that enables the mapping of words into probability densities in the latent space.

A key advantage of BHWR is the utilization of word taxonomy for the propagation of relevant information between related words. For example, consider the words 'anode' and 'cathode'. Both words have a common relationship to the word

---
* Equal contribution.

'electrode' which appears hierarchically above them in the taxonomy knowledge base. Assume the word 'cathode' frequently appears in the corpus, while the words 'anode' and 'electrode' do not appear in the corpus or occur very infrequently. A model that relies solely on co-occurrence relations will fail to infer the semantic proximity between 'cathode' and 'anode'. However, a model that utilizes word taxonomy will learn a representation for the parent word 'electrode' based on its child 'cathode'. Moreover, the parent word 'electrode' will serve as an informative prior for 'anode', and the representation of 'anode' will fall back to its prior 'electrode'. Finally, if more occurrences of 'anode' will be added to the training dataset, its representation can smoothly transition away from its prior position in accordance with the co-occurrences patterns in the data.

Besides the semantic information added to the word representations via the hierarchical prior, the Bayesian modeling by itself helps deal with the problem of rare words. These words suffer from insufficient statistics, and their respective embeddings are quite sensitive to noise. This problem becomes acute in the case of point estimate solutions that do not model uncertainty. In contrast, Bayesian solutions learn the entire posterior density and hence are more robust to overfitting (Bishop, 2006).

We train BHWR on a small annotated corpus (Miller et al., 1993) and evaluate its overall performance as well as the improvement on rare words. Our findings show that BHWR outperforms other non-contextualized word embedding methods that facilitate either Bayesian modeling or semantic taxonomy.

## 2 Related Work

Incorporating lexical-semantic information in learning word embeddings has been suggested in the past. In (Faruqui et al., 2015), a post-processing technique was introduced in order to refine pre-trained word representations using relational information from semantic lexicons. In (Li et al., 2016), hierarchical taxonomy was utilized for improving document categorization and concept clustering. Recently, linguistic knowledge bases were utilized for enhancing *contextualized* word embeddings (Huang et al., 2019; Levine et al., 2019).

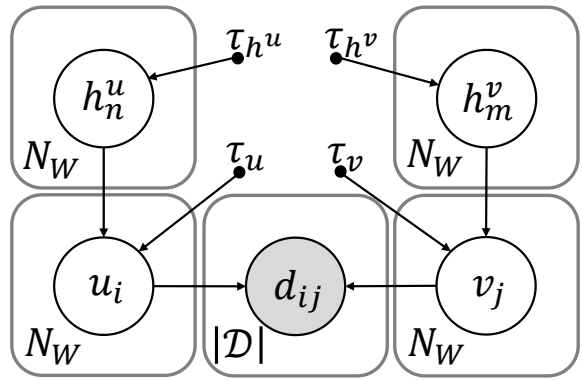

Figure 1. A graphical model of BHWR.

The abovementioned works finetune pre-trained representations, while injecting external contextual information for words (some w.r.t. a specific task). Unlike these works, BHWR facilitates Bayesian learning of non-contextualized word embeddings, in combination with hierarchical taxonomy information which is not task-specific. Hence, a direct comparison between BHWR and these works is unfitting.

More relevant to our work is the Bayesian Skip-Gram (BSG) model from (Barkan et al., 2017). However, BSG does not allow the use of external information such as word taxonomy. Hence, in our experiments, we compared BHWR to BSG and further apply the method from (Faruqui et al. 2015) to enhance BSG with word taxonomy information.

## 3 Bayesian Hierarchical Words Representation

In this section, we describe the model and derive a VB solution that is finally translated to the Bayesian Hierarchical Words Representation (BHWR) learning algorithm.

### 3.1 Model

Let $W = \{w_i\}_{i=1}^{N_W}$ and $\mathcal{I} = \{i\}_{i=1}^{N_W}$, be a vocabulary and a corresponding index set, respectively. We define $\pi: \mathcal{I} \to \mathbb{P}(\mathcal{I})$ and $\omega: \mathcal{I} \to \mathbb{P}(\mathcal{I})$ ($\mathbb{P}(\mathcal{I})$ is the power set of $\mathcal{I}$) s.t. $\pi_i \triangleq \pi[i]$ and $\omega_i \triangleq \omega[i]$ are the sets of parents and children word indices for the word index $i$, respectively. This forms a hierarchical structure (network) in which a word can appear either as a leaf or as an internal node (parent).

Let $u_i, v_i, h_i^u, h_i^v \in \mathbb{R}^k$ be the context and target leaf and parent representations of the word $w_i$,

respectively. For example, if $n \in \pi[i]$, then we use $h_n^u$ as the parent node for both $u_i$ and $h_i^u$. In addition, we define $s_i^u \triangleq |\pi_i|^{-1} \sum_{n \in \pi_i} h_n^u$ if $\pi_i \neq \emptyset$, otherwise $s_i^u \triangleq \vec{0}$, where $s_i^v$ is defined in the same manner.

Let $T = (w_{t_i})_{i=1}^L$ be a text corpus and let $c_{max} \in \mathbb{N}$ be the context window parameter. We iterate over $T$ and for each word $w_{t_i}$, we sample a random window size $c \in \{1, \ldots, c_{max}\}$ to form a multiset of positive examples $I_P^i = \{(t_i, j) | j \in \{t_{i-c}, \ldots, t_{i+c}\} \setminus \{t_i\}\}$ and a corresponding multiset of negative examples $I_N^i = \{(t_i, n_j)\}_{j=1}^{|I_P^i|}$, where $n_j$ is sampled according to the unigram distribution raised to the power of 0.75. Then, we define $I_P \triangleq \bigcup_{i=1}^L I_P^i$ and $I_N \triangleq \bigcup_{i=1}^L I_N^i$ as the positive and negative sets and further define $I_D \triangleq I_P \cup I_N$.

Let $d: \mathcal{J} \times \mathcal{J} \to \{1, -1\}$ with $d_{ij} = 1$ if $(i,j) \in I_P$ and $-1$ otherwise, and let $\mathcal{D} = \{d_{ij} | (i,j) \in I_D\}$. We model the likelihood of $d_{ij}$ given the model parameters as $p(d_{ij}|u_i, v_j) = \sigma(d_{ij} u_i^T v_j)$, where $\sigma(a) \triangleq \frac{1}{1+e^{-a}}$. We further assume normal hierarchical priors as follows:

$$p(H^u|\tau_{h^u}) = \prod_{i \in \mathcal{J}} \mathcal{N}(h_i^u; s_i^u, \tau_{h^u}^{-1} I),$$

$$p(U|H^u, \tau_u) = \prod_{i \in \mathcal{J}} \mathcal{N}(u_i; s_i^u, \tau_u^{-1} I),$$

where $\tau_{h^u}$ and $\tau_u$ are the precision hyperparameters. In the same manner, we assume normal hierarchical priors $p(V|H^v, \tau_v)$ and $p(H^v|\tau_{h^v})$. Then, the joint density of $\mathcal{D}$ and the model parameters $\theta = \{U, V, H^u, H^v\}$ given the precision hyperparameters $\mathcal{T} = \{\tau_u, \tau_v, \tau_{h^u}, \tau_{h^v}\}$ is given by

$$p(\mathcal{D}, \theta | \mathcal{T}) = p(\mathcal{D}|\theta) p(\theta | \mathcal{T}), \quad (1)$$

with

$$p(\mathcal{D}|\theta) = p(\mathcal{D}|U, V) = \prod_{(i,j) \in I_D} \sigma(d_{ij} u_i^T v_j),$$

and

$$p(\theta | \mathcal{T}) =$$
$$p(U|H^u, \tau_u) p(H^u|\tau_{h^u}) p(V|H^v, \tau_v) p(H^v|\tau_{h^v}).$$

Figure 1 presents a graphical model of BHWR.

Our goal is to compute posterior predictive distribution for an arbitrary $d_{ij}^*$ given $\mathcal{D}$ (which is not necessarily in $\mathcal{D}$). The probability of the words $w_i$ and $w_j$ to co-occur is given by

$$p(d_{ij}^* = 1 | \mathcal{D}, \mathcal{T}) = \int \sigma(u_i^T v_j) p(\theta|\mathcal{D}, \mathcal{T}) d\theta. \quad (2)$$

### 3.2 Posterior Approximation

Since the posterior $p(\theta|\mathcal{D}, \mathcal{T})$ in Eq. (2) is intractable, we turn to VB approximation (Bishop, 2006) of $p(\theta|\mathcal{D}, \mathcal{T})$ via a fully factorized distribution

$$q(\theta) = q(U) q(V) q(H^u) q(H^v)$$
$$\prod_{i \in \mathcal{J}} q(u_i) \prod_{i \in \mathcal{J}} q(v_i) \prod_{i \in \mathcal{J}} q(h_i^u) \prod_{i \in \mathcal{J}} q(h_i^v).$$

The posterior approximation $q(\theta)$ is obtained via the minimization of the KL divergence from the true posterior, namely the minimization of $D_{KL}(q(\theta)||p(\theta|\mathcal{D}, \mathcal{T}))$, which is equivalent (Bishop, 2006) to the maximization of (negative) variational free energy

$$\mathcal{L}(q) \triangleq \int q(\theta) \log \frac{p(\theta|\mathcal{D}, \mathcal{T})}{q(\theta)} d\theta.$$

$\mathcal{L}(q)$ is maximized via an iterative procedure that is guaranteed to converge to a local optima (as the optimization is non-convex): At each iteration, we update each parameter $z \in \theta$, in turn, according to the following update rule:

$$q^*(z) = \exp(\mathbb{E}_{q(\theta \setminus z)}[\log p(\theta, \mathcal{D}|\mathcal{T})] + c). \quad (3)$$

However, a straightforward application of Eq. (3) will run useless, as the term $p(\theta, \mathcal{D}|\mathcal{T})$ includes the likelihood $p(\mathcal{D}|\theta)$, which consists of sigmoid functions that are not conjugate to the normal prior $p(\theta|\mathcal{T})$ from Eq. (1). Therefore, by introducing an additional variational parameter $\xi_{ij}$, we can utilize the logistic bound from (Jaakkola and Jordan, 1996)
for lower bounding the log likelihood $p(\mathcal{D}|\theta)$ with a squared exponential function as follows:

$$\log p(\mathcal{D}|\theta) \geq \log p_\xi(\mathcal{D}|\theta) =$$
$$\sum_{(i,j) \in I_D} \frac{d_{ij}(u_i^T v_j + b_j)}{2} - \lambda(\xi_{ij})(u_i^T v_j v_j^T u_i - \xi_{ij}) + \log \sigma(\xi_{ij})$$

with $\lambda(a) \triangleq \frac{1}{2a}\left(\sigma(a) - \frac{1}{2}\right)$.

Moreover, this bound is tight for

$$\xi_{ij} = \sqrt{\sum_{m=1}^k (\sigma_{u_{ik}}^2 + \mu_{u_{ik}}^2)(\sigma_{v_{jk}}^2 + \mu_{v_{jk}}^2)}. \quad (4)$$

$p_\xi(\mathcal{D}|\theta)$ enables a conjugate relation with $p(\theta|\mathcal{T})$ that results in normal density estimators $q^*(z), z \in \theta$. Hence, for each $z \in \theta$, we update the precision $P_z$ and mean $\mu_z$ (the sufficient statistics), following the update rule from Eq. (3).

Specifically, for $q(h_i^u)$ and $q(u_i)$, the parameters updates are

$$P_{h_i^u} = \left(\tau_{h^u} + \sum_{m \in \omega_i} \frac{\tau_u + \tau_{h^u}}{|\pi_m|^2}\right) I, \quad (5)$$

$$\mu_{h_i^u} = P_{h_i^u}^{-1} \left[\sum_{m \in \omega_i} \left(\frac{1}{|\pi_m|}(\tau_u \mu_{u_m} + \tau_{h^u} \mu_{h_m^u}) - \frac{\tau_u + \tau_{h^u}}{|\pi_m|^2} \sum_{n \in \pi_m \setminus \{i\}} \mu_{h_n^u}\right)\right],$$

and

$$P_{u_i} = \tau_u I + 2 \sum_{j \in I_{u_i}} \lambda(\xi_{ij}) \mathbb{E}_{q(\theta \setminus u_i)}[v_j v_j^T], \quad (6)$$

$$\mu_{u_i} = \frac{1}{2} P_{u_i}^{-1} \left[\sum_{j \in I_{u_i}} d_{ij} \mu_{v_j} + \frac{\tau_u}{|\pi_i|} \sum_{m \in \pi_i} \mu_{h_m^u}\right],$$

respectively, where $I_{u_i} = \{j | (i,j) \in I_D\}$ and $\mathbb{E}_{q(\theta \setminus u_i)}[v_j v_j^T] = P_{u_i}^{-1} + \mu_{v_j} \mu_{v_j}^T$. The parameter updates for $q(h_i^v)$ and $q(v_i)$ are symmetric in the Eqs. (5) and (6), respectively.

### 3.3 The BHWR Algorithm

The BHWR algorithm can be summarized as follows:
1. For each $z \in \theta$, sample $\mu_z \sim \mathcal{N}(0, I)$ and initialize $P_z = I$.
2. Update $\xi_{ij}$ using Eq. (4), update $q(h_i^u)$ and $q(u_i)$ using Eqs. (5) and (6), and update $q(h_i^v)$ and $q(v_i)$ using the symmetric versions of Eq. (5) and Eq. (6), respectively.
3. Repeat step 2 until convergence.

### 3.4 Posterior Predictive Approximation

Finally, we approximate the integral from Eq. (2) by replacing the posterior with its factorized approximation

$$\int \sigma(u_i^T v_j) p(\theta | \mathcal{D}, \mathcal{T}) d\theta \approx \int \sigma(u_i^T v_j) q(\theta) d\theta$$
$$\approx \int \sigma(x) \mathcal{N}(x; \mu_x, \sigma_x^2) dx$$
$$\approx \sigma\left(\mu_x / \sqrt{1 + \pi \sigma_x^2 / 8}\right), \quad (7)$$

where $x = u_i^T v_j$ and its density is approximated using normal density (using $x$'s first two moments under $q$). The final transition follows the logistic Gaussian integral approximation suggested by (MacKay, 1992).

In practice, the similarity score for a pair of words $w_i$ and $w_j$ was based on two different versions of Eq. (7): The first by assigning $x = u_i^T u_j$ and the second with $x = v_i^T v_j$. Then, the average of these two scores is taken as the final similarity score. Our experiments revealed that this technique yields better results.

## 4 Experimental Setup and Results

The experimentations in this section are focused on word similarity. Next, we present the training corpus, evaluated models, evaluation tasks, and the results.

### 4.1 Training Corpus

We use SemCor (Miller et al., 1993), which contains 37,176 annotated sentences with 820,411 words and a vocabulary size of 11,766 words. Each word's parent is taken to be its WordNet (Miller et al., 1990) hypernym, e.g., for the words 'anode' and 'cathode', the parent word is 'electrode'.

### 4.2 Models and Configurations

We compare Bayesian Hierarchical Words Representation (**BHWR**) with the Skip-Gram with negative sampling (**SG**) model from (Mikolov et al., and the Bayesian Skip-Gram (**BSG**) model from (Barkan, 2017). For each model, we consider two versions: The first uses the word representations produced by the model as is. In the second version, we further refine the learned word representations by applying the post-processing step from (Faruqui et al., 2015). This enables the incorporation of word taxonomy information also to the SG and BSG methods. Overall, we consider six different model configurations; the post-processing versions of the modeled are marked with a '-P' suffix.

All models were trained till convergence. We used subsampling parameter (Mikolov et al., 2013) of $10^{-4}$ and a negative to positive ratio of 1. The precision hyperparameters were set to $\tau_u = \tau_v = 0.1$ and to $\tau_{h^u} = \tau_{h^v} = 0.001$. The embedding dimension was set to $k = 50$.

### 4.3 Evaluation Tasks

The word similarity evaluation includes several different datasets: WordSim-353 (**WS**) (Finkelstein et al., 2001), Stanford's Contextual Word Similarities (**SCWS**) (Huang et al., 2012), Rare Words (**RW**) (Luong et al., 2013), **MEN** (Bruni et al., 2014) and SimLex-999 (**SL**) (Hill et al., 2015). Note that these datasets are annotated by humans' similarities of words.

For BHWR and BSG, scoring a pair of words is done by using the posterior predictive approximation (Section 3.4). For SG, we compute $\frac{1}{2}(\mathcal{C}(u_i, u_j) + \mathcal{C}(v_i, v_j))$, where $\mathcal{C}$ is the cosine

| Model  | MEN  | RW   | SCWS | SL   | WS   | AVG  |
|--------|------|------|------|------|------|------|
| BHWR   | **42.7** | **28.2** | **43.2** | **15.6** | **38.2** | **33.6** |
| BSG    | 35.0 | 27.6 | 39.6 | 13.2 | 28.6 | 28.8 |
| SG     | 38.3 | 23.0 | 36.2 | 13.6 | 27.0 | 27.6 |
| BHWR-P | **49.9** | **29.2** | **47.6** | **18.5** | **38.6** | **36.8** |
| BSG-P  | 40.6 | 28.2 | 41.9 | 14.6 | 31.5 | 31.4 |
| SG-P   | 41.2 | 27.7 | 41.0 | 14.7 | 31.2 | 31.2 |

Table 1: Word similarity evaluation.

| Model  | MEN  | RW   | SCWS | SL   | WS   | AVG  |
|--------|------|------|------|------|------|------|
| BHWR   | **37.7** | **25.3** | **31.7** | **9.9** | **42.9** | **29.5** |
| BSG    | 27.9 | 24.9 | 27.4 | 9.4 | 34.4 | 24.8 |
| SG     | 28.2 | 24.2 | 23.1 | 9.2 | 22.5 | 21.4 |
| BHWR-P | **46.4** | **29.2** | **35.9** | **12.2** | **42.8** | **33.4** |
| BSG-P  | 37.2 | 25.6 | 30.3 | 11.0 | 32.0 | 27.2 |
| SG-P   | 37.0 | 25.4 | 30.6 | 11.2 | 30.8 | 27.0 |

Table 2: Word similarity evaluation for rare words.

similarity function (recall SG is based on a point estimate solution).

Finally, for each combination of dataset and method, we report the Spearman rank correlation in terms of percentage.

### 4.4 Results

Table 1 presents the results for all combinations of models and datasets. In the last column, we report for each model the average score across all datasets. The table is partitioned into two sections that present the regular and the post-processed versions of the models. For each dataset and section, the best and second-best scores are boldfaced and underlined, respectively. Next, we turn to discuss the main trends presented in Tab. 1.

First, we consider the regular model versions (first three rows). BHWR significantly outperforms BSG and SG across all datasets, and BSG comes second with a noticeable difference. This demonstrates the merit of the Bayesian treatment (BSG ≻ SG) and the modeling of word taxonomy (BHWR ≻ BSG).

Next, we turn to examine the post-processed versions (last three rows). We observe a significant boost to the results of all the models, which serves as an independent evaluation and reinforcement to the effectiveness of the post-processing method from (Faruqui et al., 2015). BHWR-P again surpasses the other models by a large margin, while BSG-P and SG-P are on par. An interesting observation is that the post-processing method is found to be instrumental not only for SG and BSG but also for BHWR that utilizes word taxonomy inherently (BHWR ≺ BHWR-P). This can be explained by the fact that the method of (Faruqui et al., 2015) uses additional lexical information such as synonyms, which are not incorporated in BHWR. Yet, BHWR alone (without post-processing) still outperforms both BSG-P and SG-P. This result demonstrates the advantage of BHWR that facilitates learning of co-occurrences relations together word taxonomy, simultaneously.

Note that the results in Tab.1 are suboptimal when compared to (Pennington et al., 2014): This is clearly related to the small corpus size used in this work. In the future, we plan to conduct an evaluation on larger corpora that are not necessarily annotated.

Finally, In order to demonstrate the strength of BHWR for words with only a few occurrences in the corpus, we further compare the models' performance on rare words. Table 2 shows the results on the word similarity tasks for words that occurred in the corpus five times or less. We observe that the gaps between BHWR and the other models become even more significant, either with or without the utilization of the post-processing from (Faruqui et al., 2015).

## 5 Conclusion and Future Work

We presented BHWR - A word representation learning model, facilitating Bayesian learning of co-occurrences relations together with word taxonomy via hierarchical priors. When trained on a small corpus, BHWR exhibits a significant performance gain over other word embedding methods across various word similarity datasets. Importantly, a remarkable improvement is obtained for rare words. Moreover, BHWR outperforms all other baselines even when the latter are enhanced with the post-processing taxonomy refinement procedure from (Faruqui et al., 2015). Finally, when combining BHWR with the post-processing from (Faruqui et al., 2015), further improvement is observed.

In the future, we plan to extend the applicability of the presented model to other linguistics tasks as well as recommendations and medical inference tasks.